\documentclass[letterpaper, 10 pt, conference]{format/ieeeconf}

\usepackage{amsmath}
\usepackage{amsfonts}

\usepackage{amsthm}
\usepackage[ruled,vlined]{algorithm2e}
\usepackage{mathtools}
\usepackage{tabularx}
\usepackage{graphicx}
\usepackage{subcaption} 
\usepackage{enumerate}
\usepackage{booktabs}
\usepackage{float}
\usepackage{url}
\usepackage{verbatim}
\usepackage{booktabs} 
\usepackage{multirow}
\usepackage{color}
\usepackage{pifont}
\usepackage[linkcolor=black,citecolor=black,urlcolor=black,colorlinks=true]{hyperref}
\usepackage{cite}

\graphicspath{{figures/}}
\IEEEoverridecommandlockouts
\title{Towards Efficient Trajectory Generation\\ for Ground Robots beyond 2D Environment}

\author{Jingping Wang$^{*1,2}$, Long Xu$^{*1,2}$, Haoran Fu$^{3}$, Zehui Meng$^{4}$,\\  Chao Xu$^{1,2}$, Yanjun Cao$^{1,2}$, Ximin Lyu$^{3}$ and Fei Gao$^{1,2}$
    \thanks{This work is supported by the Technology Cooperation Project (TC20220507025) from the Application Innovation Laboratory, Huawei Technologies Co., Itd.}
	\thanks{*Indicates equal contribution.}
	\thanks{$^{1}$State Key Laboratory of Industrial Control Technology, Zhejiang University, Hangzhou 310027, China. (\textit{Corresponding author: Fei Gao})}
	\thanks{$^{2}$Huzhou Institute of Zhejiang University, Huzhou 313000, China.}
	\thanks{$^{3}$School of Intelligent Systems Engineering, Sun Yat-sen University, Shenzhen 518107, China.}
	\thanks{$^{4}$Applicaion Innovation Laboratory, Huawei Technologies Co., Ltd.}
	\thanks{E-mail: {\tt\small \{gaolon, cxu, and fgaoaa\}@zju.edu.cn}}
}

\begin{document}

    \maketitle
    \thispagestyle{empty}
    \pagestyle{empty}

\begin{abstract}
With the development of robotics, ground robots are no longer limited to planar motion. 
Passive height variation due to complex terrain and active height control provided by special structures on robots require a more general navigation planning framework beyond 2D.
Existing methods rarely considers both simultaneously, limiting the capabilities and applications of ground robots.
In this paper, we proposed an optimization-based planning framework for ground robots considering both active and passive height changes on the z-axis.
The proposed planner first constructs a penalty field for chassis motion constraints defined in $\mathbb{R}^3$ such that the optimal solution space of the trajectory is continuous, resulting in a high-quality smooth chassis trajectory.
Also, by constructing custom constraints in the z-axis direction, it is possible to plan trajectories for different types of ground robots which have z-axis degree of freedom. We performed simulations and real-world experiments to verify the efficiency and trajectory quality of our algorithm.

\end{abstract}

\section{Introduction}
\label{sec:Introduction}

\begin{figure}	
		\centering
		\includegraphics[width=1.0\columnwidth]{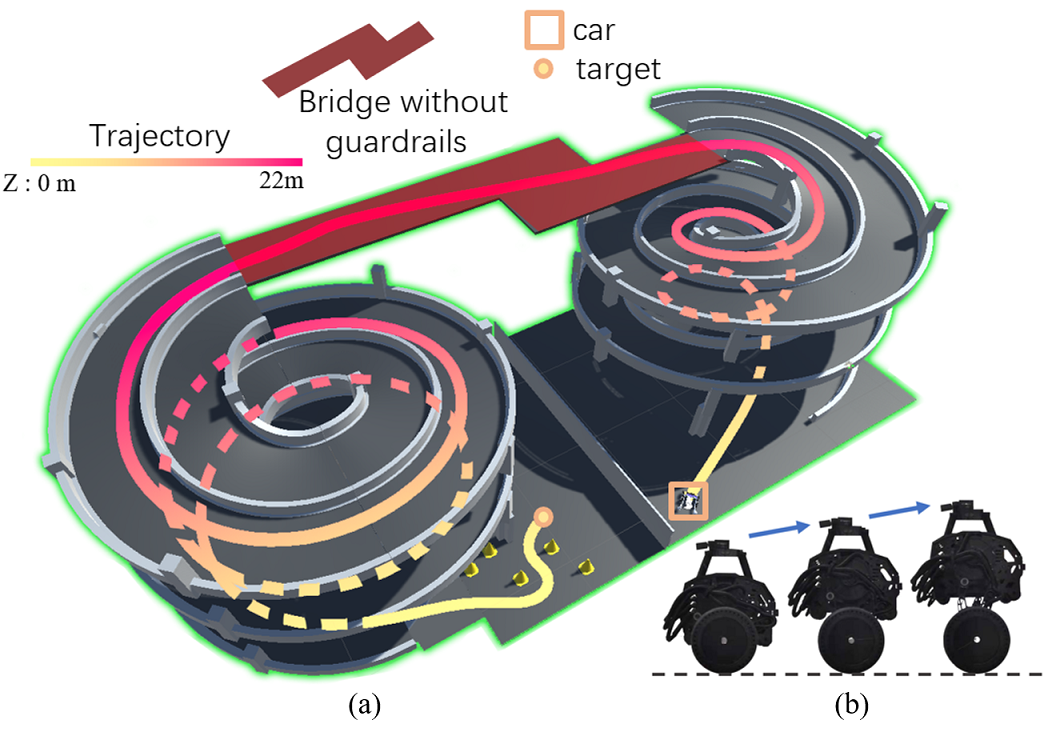}
		\caption{Two cases where the z-axis height of the ground robot changes.
		In Fig.(a), the robot must pass through a complex overpass to reach the target, 
		and the height of the chassis is passively varied by terrain constraints.
        Fig.(b) shows a wheeled bipedal robot that can change the height of its head to accomplish more complex tasks.}\label{fig:flag}	
		\vspace{-0.3cm}
\end{figure}

Navigation is an increasingly important task with applications in autonomous systems such as robotics and autonomous driving.
Among different types of robots, ground robots are seemed to doomed to be constrained on the ground, 
where the terrain shape should be taken into account. 
Furthermore, with the advancement in mechanism, sensing, and designs, the application scenarios for ground robots continue to be enriched. Many ground robots are now capable to work beyond a flat plane. 
Specifically, there are two situations shown in Fig.\ref{fig:flag} where it is necessary to consider the change in the height of the robot.
\begin{itemize}
    \item [1)]The ground is not flat, it may be rugged terrain or facilities with multi-layered structures.
    In this case, the ground robot's chassis height will passively undulate with the terrain (Fig.\ref{fig:flag}a).
	\item[2)]The robot has degrees of freedom perpendicular to the horizontal direction, such as wheeled bipedal robots\cite{klemm2019ascento,zhang2019system} and mobile manipulator robots\cite{mobileM1,mobileM2}.
	In this case, the ground robot can actively change the height of certain parts (Fig.\ref{fig:flag}b).
\end{itemize}
However, dealing with the above situations in autonomous navigation is non-trivial. Traditional works\cite{urmson2008autonomous,ziegler2009spatiotemporal,rufli2010design,mcnaughton2011motion,han2011unified,palmieri2016rrt,lavalle1998rapidly,karaman2011sampling,webb2013kinodynamic,werling2010optimal,ding2019safe,ding2021epsilon} relieve this by assuming that the robot work on a flat surface, which largely simplifies the robot’s work scenarios and restricts the motion capabilities. Some  works\cite{krusi2017driving,liu2015robotic,stumm2012tensor} consider complex terrain but ignore the active height change of ground robots. The inability to well handle the passive and active height variation will greatly limit the capabilities of ground robots, further limiting the applications.

%However, existing works do not handle the z-axis trajectory of ground robots perfectly.

To compensate for the shortcomings of existing methods, in this paper, we design an optimization-based planning framework for ground robots to generate trajectories considering both active and passive height changes. 

The proposed algorithm is composed of a pre-analysis procedure and a gradient-based Spatio-temporal trajectory optimizer. 
Firstly, we filter the original point cloud to provide a search space for the front end of the trajectory generation considering the robot's size. We then construct a penalty field defined in $\mathbb{R}^3$ which embodies the characteristics of the terrain shape. Finally, we optimize the trajectory with smoothness, collision, dynamical feasibility, terrain traversability, and other customized cost terms. Our algorithm can cope with non-planar environments and supports active z-axis degree of freedom.
We perform comprehensive tests in simulation and the real world to validate our method. Contributions of this paper are:

\begin{itemize}
\item [1)]We propose a novel field construction algorithm in $\mathbb{R}^3$, which describes the level of danger for a robot trying to get a foothold in space.%which evaluates the ground shape and be defined in $\mathbb{R}^3$.
\item [2)]We propose an optimization-based planning framework for ground robots, which can generate trajectories beyond 2D space efficiently.
\item [3)]We open source our software\footnote{\url{https://github.com/ZJU-FAST-Lab/3D2M-planner}} for the reference of the community.

\end{itemize}

\section{Related Works}
\label{sec:RelatedWork}

\subsection{Navigation in 2D Cartesian space}
\label{subsec:Navigation in 2D space}
Navigation in 2D Cartesian space for ground robots can be translated into elegant mathematical problems, spawning numerous practical studies.
Methods can be divided into sampling-based methods and optimization-based methods.
Sampling-based methods \cite{urmson2008autonomous,ziegler2009spatiotemporal,rufli2010design,mcnaughton2011motion} are widely used for robot motion planning.
The probabilistic planners \cite{han2011unified,palmieri2016rrt} represented by Rapidly-exploring Random Tree (RRT) \cite{lavalle1998rapidly} obtain a feasible path by expanding the state tree rooted at the starting node. 
Some improved approaches \cite{karaman2011sampling,webb2013kinodynamic} were later proposed.
However, sampling-based methods confront a dilemma between computation consumption and trajectory quality which limits the direct application in realistic settings.
Optimization-based methods \cite{werling2010optimal,ding2019safe,ding2021epsilon} usually construct the planning problem as a mathematical optimization problem and solved using numerical optimization algorithms.
Commonly, the safety constraints are made by safe motion corridor or Euclidean Signed Distance Functions (ESDF)\cite{esdf}. %\textbf{S}afe \textbf{M}otion \textbf{C}orridor(SMC) or \textbf{E}uclidean \textbf{S}igned \textbf{D}istance \textbf{F}unctions(ESDF)\cite{esdf}.

\subsection{Navigation considering the terrain shape}
\label{subsec:Navigation on point clouds}
Trajectories computed with the assumption of flat terrain may turn out to be highly suboptimal or even unfeasible when mapped onto the 3D terrain surface.
To incorporate the 3D shape of the terrain in motion planning of ground robots,
several works have been proposed.

Kr{\"u}si et al. presented a practical approach\cite{krusi2017driving} to global motion planning and terrain assessment for ground robots in generic 3D environments, which assesses terrain geometry and traversability on demand during motion planning. It consists of sampling-based initial trajectory generation, followed by precise local optimization according to a custom cost measure to directly generate a feasible path connecting two states in the state space of a ground robot's chassis.
However, simultaneous terrain assessment during path searching seriously reduces computational efficiency, and the initial RRT-Connect\cite{rrt-connect} trajectory sampling results in long path search times when dealing with large or complex scenes. 
Additionally, the final trajectory is biased to longer due to no prior global evaluation.

To solve the above problems, some approaches perform pre-evaluation of global point clouds based on tensor voting\cite{medioni2000tensor}, improving planning efficiency and trajectory quality.
\cite{stumm2012tensor} described a solution to robot navigation on 3D surfaces, which utilized an improved A* algorithm for path searching, significantly increasing efficiency. But this approach is only applicable to special robots with adsorption ability.\cite{liu2015robotic} uses sparse tensor voting and accelerates the process with GPU. After building a connectivity graph between global point clouds by the concept of k-Nearest-Neighbor (k-NN) and constructing a Riemannian metric, it searches for trajectories towards the flattest direction. Nevertheless, due to the lack of enlightenment, the Dijkstra algorithm it uses is inefficient.

\section{Point Cloud Pre-processing}
\label{sec:Point Cloud Pre-processing}
Inflating point clouds is a typical approach in motion planning to ensure safety.
However, this approach is not suitable for our work, because some point clouds in the environment represent traverasable areas like the ground, and cannot be simply considered as obstacles.
In fact, in the process of optimization-based trajectory generation, 
the final trajectory must be safer than the initial trajectory after the 
optimization process due to the safety constraints. 
Therefore, the key is to ensure that the initial trajectory is in the robot‘s C-space. %FF
In this work, the initial trajectory is provided by a front-end path searcher.

To this end, we proposed a more concise approach to ensure the safety of the initial trajectory.
This approach directly processes the original point cloud and is named  Valid Ground Filter (VGF).
The raw point cloud records the spatial location information of the object surface, but not all parts of the object surface are available for the robot to stand on.
We defined criteria for the ``valid'' points in the point cloud as follows:
\begin{itemize}
    \item [1)]The point is on the upper surface of an object.
    
    \item [2)]The local area near the point is flat enough which means the robot does not collide or topple.%FF
    
\end{itemize}
Denote the original point cloud by $ M $ and the point cloud after VGF by $ M^s $. $ M^s $ is defined as follow:

\begin{align}
\begin{split}
M^s = & \{ p \in M | D = K(p,r) , \forall p^d \in D, p^d_z \le p_z \ or \\
p^d_z > p_z ,
    & \arctan{(\frac{p_z - p^d_z}{\sqrt{(p_x - p^d_x)^2+( p_y - p^d_y)^2}})} < \theta  ) \},
\end{split}
\end{align}
where $D$ is the neighboring points of $p$ within radius $r$. In practice, the value of $r$ should be greater than the minimum ball radius that can wrap around the robot, and $ \theta $ is the maximum tilt angle of the surface on which the robot can stand. 
The pipeline for iteratively processing each point is stated in Algorithm \ref{alg:Point Cloud Pre-processing}.

\begin{figure}	
		\centering
		\includegraphics[width=1.0\columnwidth]{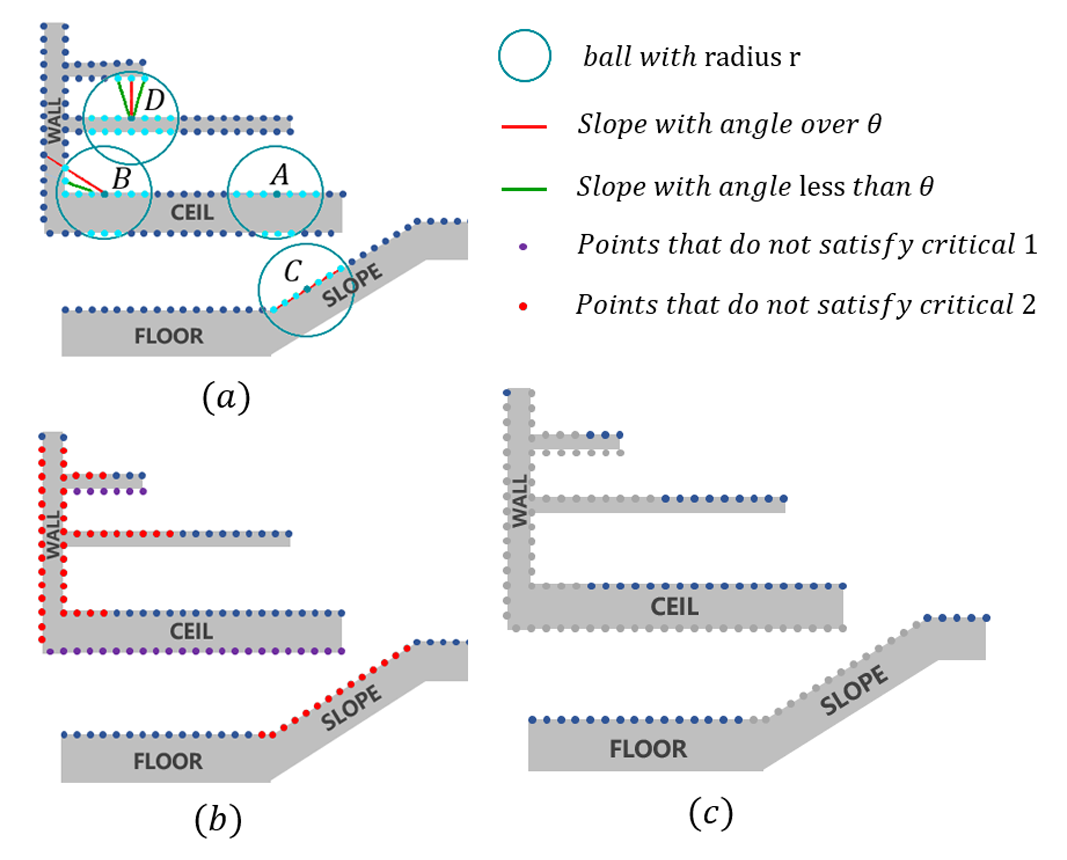}
		\caption{The Valid Ground Filter Processing. To facilitate this we show side views. Figure (a) illustrates the specific process for four points:
		        According to the criteria, point A is feasible. Point B is too close to the lateral obstacle.
		        Point D is too close to the obstacle above. And the plane where point C is located is too inclined.
		        Figure (b) shows the determination of the points after completing a traversal. Figure (c) shows the final results of VGF.}	
		\vspace{0.3cm}
		\label{fig:preprocess}
\end{figure}

Fig.\ref{fig:preprocess} illustrates the process of VGF and the result in a demo scenario. 
After VGF, the points in $ M^s $ all
represent the valid surface on which the robot can stand.
The path searching process will be performed on $ M^s $ to obtain a safe path considering the size of the robot.

\begin{algorithm}
    \caption{Valid Ground Filter}
    \label{alg:Point Cloud Pre-processing}
    \KwIn{$M,r,\theta$}
    \KwOut{$M^s$}
    \Begin
    {
        $M^s \leftarrow \emptyset$\;
        \For {\textbf{each} $p \in M$ }
        {
        
            $D_p = GET\_NEIGHBORS(p,r)$\;
            \For{\textbf{each} $p^d \in D_p $}
            {
                \uIf{$ p^d_z < p_z $}
                {
                    
                    $ \alpha = \arctan{(\frac{p_z - p^d_z}{\sqrt{(p_x - p_x^d)^2+( p_y -p_y^d)^2}})} $\;
                    
                    \uIf{$\alpha < \theta$}
                    {
                    
                        $M^s \gets M^s \cup p $\;
                    }
                }
            }  
        }
        \Return{$M^s$};
    }
\end{algorithm}

\section{Safety Penalty Field}
\label{sec:safety constraints}

For aerial robots, safety constraints can easily be added to trajectory optimization using ESDF\cite{zhou2019robust} or safe flight corridors\cite{gao2020teach} so that trajectories do not collide with obstacles.
However, since the ground robot is constrained on the ground, it is not enough for safety constraints to contain only collision-free term (for example, a trajectory that makes a vehicle run off a cliff meets the collision-free constraint but is extremely dangerous).

In this paper, the safety constraint ensures that no hazards occur while the ground robot is walking along the trajectory, 
and we define hazards as follows:
\begin{itemize}
\item [1)]Collision between robot and object.

\item [2)]The robot standing in a too tilted position.

\item [3)]The robot undergoes large positional changes (e.g. falling from the edge of a step and drastic change in pose).
\end{itemize}

To avoid situation 1), we also use the ESDF map defined in the grid map which pushes a 3-dimensional trajectory away from objects in any direction.
Moreover, the penalty function based on ESDF is defined in $\mathbb{R}^3$, 
which is convenient for the optimization problem.

To avoid situations 2) and 3), the estimation of robot posture is necessary.
We evaluated the posture of the robot in various places by fitting the local plane using RANSAC\cite{fischler1981random}. 
Specifically, estimates were made in the position of each 3D grid, considering that the penalty function should be defined in $\mathbb{R}^3$.
However, point clouds exist only close to the object. So we use a \textit{Downward Projection Strategy}:

$ p = [x,y,z]^T $ is a point in space, and $ H:\mathbb{R}^3 \mapsto \mathbb{R}^3 \times SO(2) $ 
represents the result of the robot pose estimation at position $  p $.
We define $ p^{'} = [x,y,z^{'}]^T$ the point that is projected downward onto the surface of the nearest object from $ p $.
\begin{align}
    H(p) = H(p^{'}) = [p_{c}, \phi ]^{T}
\end{align}
%$$ H(p) = H(p^{'}) = [p_{c}, \phi ]^{T} $$

The value of $H$ is the RANSAC fit to the local point cloud at $p^{'}$. 
Where $p_{c} = [x_{c}, y_{c}, z_{c}]^{T} $ is the COM(center of mass) of the local plane, 
$\phi$ is the angle between the local plane normal vector and $ [0,0,1]^{T} $. 
To avoid the errors caused by the wrong direction of normal vector, $ \phi = min(\phi, \pi - \phi)$.

Obviously, for a given position $p$ , a larger $\phi$ is more detrimental to the robot.
And 2-norm of the gradient $|| \nabla H(p) ||_{2}$ reflects the sharpness of the robot's positional change when moving near $p$.
Therefore, we defined a travel penalty function $S:\mathbb{R}^3 \mapsto \mathbb{R}$:
%  S(p) = \lambda_{f} H^{2}_{(4)}(p) + \lambda_{m} ( \frac{\partial H(p)}{\partial x} + \frac{\partial H(p)}{\partial y} + \frac{\partial H(p)}{\partial z} + \frac{\partial H(p)}{\partial \phi}), %\nabla ), %\partial H(p) ),
\begin{align}
S(p) = \lambda_{f} H^{2}_{(4)}(p) + \lambda_{m} (\sqrt{|| \nabla H(p) ||_{2}} ),
\end{align}
where $\lambda_{f}$ and  $\lambda_{m}$ are weights.

Thanks to the downward projection strategy, the travel penalty function $S$ is defined in the entire $\mathbb{R}^3$ but there are only horizontal gradients existing on a continuous terrain. In addition, 
to ensure a certain safety margin, we performed equidistant diffusion of function $S$, denoted by $S'$.
Since the map is represented by a grid-map, the diffusion process also proceeds discretely.
Assume that $t$ is the diffusion distance, construct the matrix $A \in \mathbb{R}^{m \times m} $ which meets the following conditions:

\begin{itemize}
\item [1)] $m = 2 \lceil t/r_{g} \rceil + 1$ , where $r_{g} $ is the resolution of the gridmap.
\item [2)] $A_{i,j} = \max\{ 0, 1 - r_{g}\frac{\sqrt{|i - \frac{m-1}{2}|^{2} + |j - \frac{m-1}{2}|^{2} }}{t} \} $.
\end{itemize}

Then, we can get the $S'$ by performing a convolution operation on the values of the function $S$ at each horizontal layer with $A$ as the convolution kernel. 

\begin{figure}
        \vspace{-0.4cm}
		\centering
		\subcaptionbox{the point cloud\label{fig:ori_pc}}{
			\includegraphics[width=0.35\columnwidth]{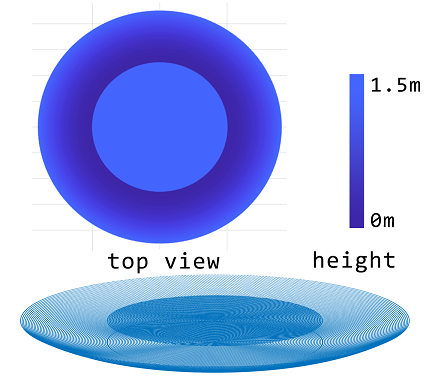}}
		\quad
		\subcaptionbox{top view}{
			\includegraphics[width=0.45\columnwidth]{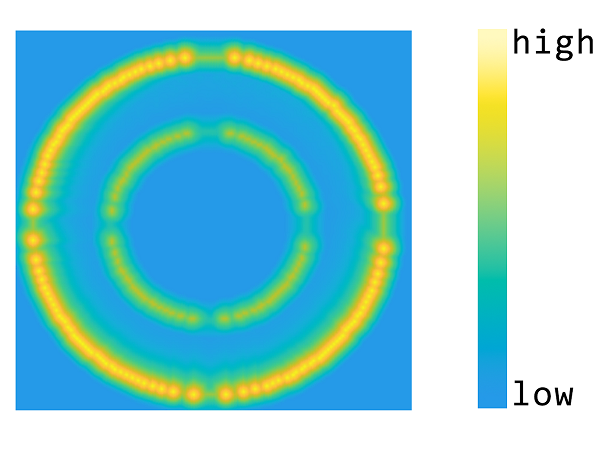}}
		\quad
		\subcaptionbox{side view}{
			\includegraphics[width=0.45\columnwidth]{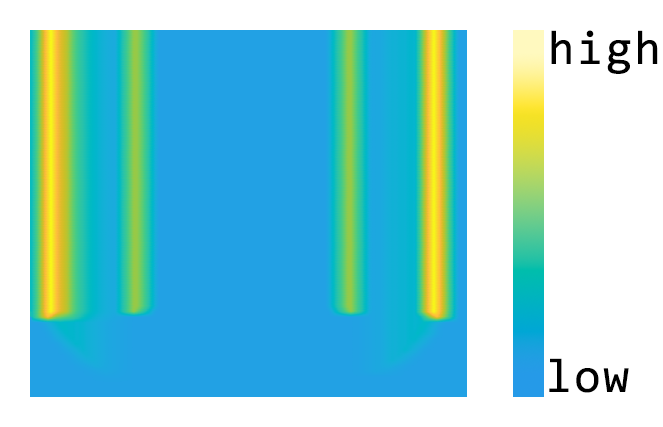}}
		\caption{Distribution of function $S$ in a demonstrative environment, which describes some of the properties of the safety penalty field proposed in Sec.\ref{sec:safety constraints}. Fig.(a) shows the point cloud map of the example environment 
		which includes multi-level structures and sloping floors. Fig.(b) and Fig.(c) show the distribution of the diffused function $S$
		in $\mathbb{R}^3$ space from different perspectives. Color represents the value of $S$.
		}
		\label{fig:travel_penalty}
		\vspace{-0.4cm}
\end{figure}

Fig.\ref{fig:travel_penalty} shows the result of function $S$ after distribution in a demonstrative environment.

\section{Trajectory Generation}
\label{sec:Trajectory Generation}
In this section, we present our method to incorporate safety and dynamical feasibility constraints into our optimization process.
In addition, custom tasks for different types of ground robots can be achieved by constructing custom constraint terms.
To begin with, we introduce a computationally efficient feasibility check method that applies to a wide range of constraints.
Then this method is used in a constrained trajectory optimization process.

\subsection{Path Searching}
\label{subsec:Path Searching}

We use a front-end path searcher to provide the initial trajectory.
Since smoothness and dynamic feasibility are considered in the back-end optimization,
the path searching becomes relatively simple and only the variation of the robot's position needs to be considered.

In this work, we use A*\cite{1968A} algorithm to search the path on the grid-map. Thanks to the VGF process, 
unsafe areas after considering the robot size have been excluded from the search space. 
%Also, the fact that robot motion is constrained on the ground surface reduces the number of cells to be searched, and only the cells in extremely close of the upper surface need to be considered. 
In addition, the fact that the robot motion is constrained on the ground surface reduces the number of cells to be searched, and only the cells extremely close to the upper surface need to be considered.

\subsection{Trajectory Representation}
\label{subsec:Trajectory Representation}

In this work, we adopt $\mathfrak{T}_{\textbf{MINCO}}$\cite{wang2022geometrically} for our trajectory representation, which is a minimum control effort polynomial trajectory class defined as
\begin{align}
\begin{split}
\mathfrak{T}_{\textbf{MINCO}}=\{p(t):[0,T_\Sigma]\rightarrow\mathbb{R}^m|\textbf{c}=\mathcal{M}(\textbf{q},\textbf{T}),\\
\textbf{q}\in\mathbb{R}^{m(M-1)},\textbf{T}\in\mathbb{R}^M_{>0}\},
\end{split}
\end{align}
where $p(t)$ is an $m$-dimensional $M$-piece polynomial trajectory with degree $N = 2s-1$, $s$ is the order of the relevant integrator chain, $\textbf{c}=(\textbf{c}^T_1,...,\textbf{c}^T_M)^T\in\mathbb{R}^{2Ms\times m}$is the polynomial coefficient, $\textbf{q} =(\textbf{q}_1,...,\textbf{q}_{M-1})$ is the intermediate waypoints and $\textbf{T}=(T_1,T_2,...,T_M)^T$ is the time allocated for all pieces, $\mathcal{M}(\textbf{q},\textbf{T})$ is the parameter mapping constructed from Theorem 2 in \cite{wang2022geometrically}, and $T_\Sigma=\sum_{i=1}^MT_i$ is the total time.

An $m$-dimensional $M$-piece trajectory is defined as
\begin{align}
p(t)=p_i(t-t_{i-1})\quad \forall t\in[t_{i-1},t_i),
\end{align}
and the $i^{th}$ piece trajectory is represented by a $N=5$ degree polynomial:
\begin{align}
p_i(t)=\textbf{c}^T_i\beta(t)\quad\forall t\in[0,T_i),
\end{align}
where $\textbf{c}_i\in\mathbb{R}^{(N+1)\times m}$ is the coefficient matrix, $\beta(t)=[1,t,...,t^N]^T$ is the natural basis, and $T_i=t_i-t_{i-1}$ is the time allocation for the $i^{th}$ piece.
$\mathfrak{T}_{\textbf{MINCO}}$ is uniquely determined by $(\textbf{q},\textbf{T})$. And the parameter mapping $\textbf{c}=\mathcal{M}(\textbf{q},\textbf{T})$ can convert trajectory representations $(\textbf{c},\textbf{T})$ to $(\textbf{q},\textbf{T})$, which allows any second-order continuous cost function $J(\textbf{c},\textbf{T})$ to be represented by $H(\textbf{q},\textbf{T})=J(\mathcal{M(\textbf{q},\textbf{T})},\textbf{T})$. Hence, we can get  $\partial H/\partial \textbf{q}$ and $\partial H/\partial \textbf{T}$ from $\partial J /\partial \textbf{c}$ and $\partial J /\partial \textbf{T}$ easily.

To cope with the constraints $\mathcal{G}(p(t),...,p^{(3)}(t))\preceq\textbf{0}$, such as safety assurance and dynamical feasibility, we discretize each piece of the trajectory as $\kappa_i$  \textit{Constraint Points}\cite{wang2022geometrically} $\tilde{p}_{i,j}=p_i((j / \kappa_i)\cdot T_i),j=0,1,...,\kappa_i-1$. Then the discrete constraints can be transformed into weighted penalty terms, which 
can form the cost function $J(\textbf{c},\textbf{T})$ by addition.

\subsection{Optimization Problem Formulation}
\label{subsec:Optimization Problem Formulation}

We formulate the beyond 2D trajectory generation for ground robots as an unconstrained optimization problem:
\begin{align}
\min_{\textbf{c},\textbf{T}} J(\textbf{c},\textbf{T})=\lambda_s J_s+\lambda_t J_t+\lambda_m J_m+\lambda_d J_d,
\end{align}
where $\lambda_k,k=\{s,t,m,d\}$ are the weights of each cost function.

\subsubsection{Safety Assurance $J_s$}
\label{subsubsecsafety Assurance}
We use the method proposed in \cite{zhou2019robust} to maintain an ESDF from point clouds. Combining it with the safety penalty field we designed in Sec.\ref{sec:safety constraints}, we can keep the robot moving in the traversable area, the cost function and gradients are:

\begin{align}
&\mathcal{G}_s(\tilde{p}_{i,j})=
\begin{cases}
0, & \text{$ S'(\tilde{p}_{i,j})\le s_{thr} $},\\
S'(\tilde{p}_{i,j})-s_{thr}, & \text{$ S'(\tilde{p}_{i,j})>s_{thr} $},\\
\end{cases}\\
&\mathcal{G}_c(\tilde{p}_{i,j})=
\begin{cases}
d_{thr}-d(\tilde{p}_{i,j}), & \text{$ d(\tilde{p}_{i,j})<d_{thr} $},\\
0,& \text{$ d(\tilde{p}_{i,j})\ge d_{thr} $},\\
\end{cases}\\
&J_s=\sum_{i=1}^M\sum_{j=0}^{\kappa_i-1}\mathcal{C}(\mathcal{G}_n(\tilde{p}_{i,j}))\cdot\frac{T_i}{\kappa_i},\\
&\frac{\partial J_s}{\partial\textbf{c}_i}=3\sum_{j=0}^{\kappa_i-1}\mathcal{Q}(\mathcal{G}_n(\tilde{p}_{i,j}))\cdot\beta(\frac{j}{\kappa_i}T_i)\cdot(\nabla S_d^T)\cdot \frac{T_i}{\kappa_i},\\
&\frac{\partial J_s}{\partial T_i}=\sum_{j=0}^{\kappa_i-1}\frac{\mathcal{Q}(\mathcal{G}_n(\tilde{p}_{i,j}))}{\kappa_i}\cdot[\mathcal{G}_n(\tilde{p}_{i,j})+\frac{3j}{\kappa_i}\cdot T_i\cdot (\nabla S_d^T) \tilde{\textbf{v}}_{i,j}],
\end{align}
where $\mathcal{G}_n=\mathcal{G}_s+\mathcal{G}_d$, $\nabla S_d^T=\nabla S'^T-\nabla d^T$, $s_{thr}$ is the safety threshold, $\nabla S'$ is the gradient of $S'$ in $\tilde{p}_{i,j}$, $d_{thr}$ is the collision threshold, $d(\tilde{p}_{i,j})$ is the distance between $\tilde{p}_{i,j}$ and the closet obstacle, $\nabla d$ is the gradient of ESDF in $\tilde{p}_{i,j}$. $\mathcal{C}(\cdot)=\max\{\cdot,0\}^3$ is the cubic penalty, $\mathcal{Q}(\cdot)=\max\{\cdot,0\}^2$ is the quadratic penalty.

\subsubsection{Total Time $J_t$}
\label{subsubsec:Total Time}
We minimize the total time $J_t=\sum^M_{i=1}T_i$ to improve the aggressiveness of the trajectory, the gradients are $\partial J_t / \partial\textbf{c}=\textbf{0}$ and $\partial J_t / \partial\textbf{T}=\textbf{1}$.

\subsubsection{Smoothness Cost $J_m$, Dynamical Feasibility $J_dtotal$}
\label{subsubsec:Smoothness Cost J_m and Dynamical Feasibility J_d}
To ensure that the trajectory is smooth enough and can be executed by the agent, we choose the third order of the $i^{th}$ piece trajectory (jerk) as control input, optimize the integration of it and limit the maximum value of velocity and acceleration in the same way as in \cite{han2021fast}. Readers can refer to \cite{han2021fast} for more details about the cost function and gradients.

\section{Results}
\label{sec:Results}

% \begin{figure}	
% 		\centering
% 		\includegraphics[width=0.9\columnwidth]{diablo.png}
% 		\caption{Hardware settings of Diablo}\label{fig:diablo}	
% 		\vspace{0.3cm}
% \end{figure}

\subsection{Implementation details}
\label{subsec:Implementation details}
To validate the performance of our method in real-world applications, we deploy it on a direct-drive self-balancing wheeled bipedal robot called Diablo\footnote{\url{https://diablo.directdrive.com/}}(Fig.\ref{fig:flag}b). All computations are performed by an onboard computer with an Intel Core i7-8550U CPU, which is shown in the right side of Fig.\ref{fig:indoor}. We utilize Fast-lio2\cite{xu2022fast} for robust LiDAR-based localization and point cloud map generation. Furthermore, a MPC controller\cite{muske1993model} with position and velocity feedback is fitted to the robot for trajectory tracking. The unconstrained optimization problem is solved by an open-source library LBFGS-Lite\footnote{\url{https://github.com/ZJU-FAST-Lab/LBFGS-Lite}}.
It is worth noting that the height of Diablo $h_{d}$ can be varied in the interval $[h_{min}, h_{max}]$, as shown in Fig.\ref{fig:flag}b. We choose $h=h_{min}$ in point cloud pre-processing and add a custom penalty to constrain this additional degree of freedom:
\begin{align}
&J_h=\sum_{i=1}^M\int_0^{T_i}||p_{i(3)} - h_{G}(p_{i}) - h_{S}(p_{i})||^2dt,
\end{align}
where $h_{G}$ is the height of ground and 
$h_{S}$ is the most suitable height calculated according to the floor height and ceiling height. 

In simulation, a desktop PC with an Intel i7-12700 CPU and Nvidia GeForce RTX2060 GPU is used.

\subsection{Real-World Experiments }
\label{subsec:Real-World Experiments}
\begin{figure}
	\centering
	\subcaptionbox{Indoor Experiment and Hardware Settings of Diablo: Diablo crouched down and passed under two tables marked as pink dot grid, the lower line is the result of the projection of the trajectory (upper line) onto the ground. The trajectory length is $6.9m$, planning time consuming is $10.1ms$. \label{fig:indoor}}{
		\includegraphics[width=1.0\columnwidth]{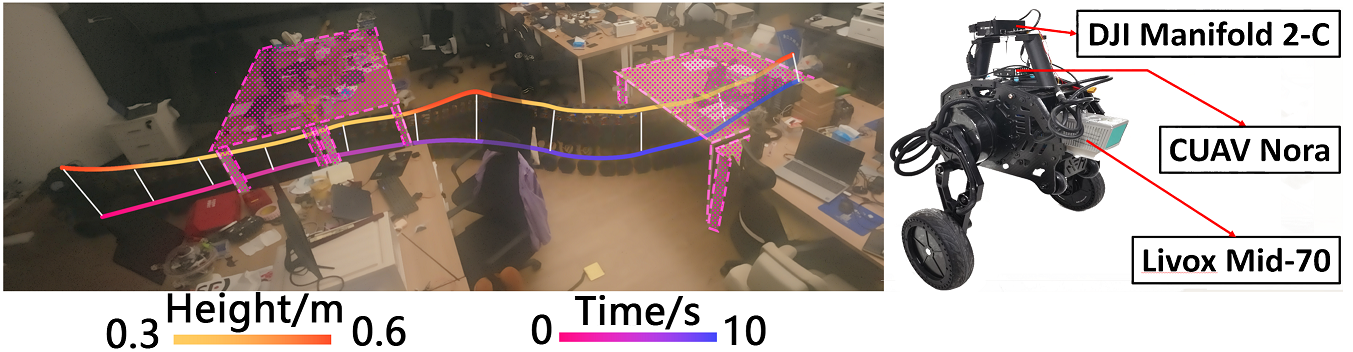}}
	\newline
	\vspace{0.3cm}
	\subcaptionbox{Outdoor Experiment: Diablo is demanded to come up to a higher plane via the barrier-free access, the trajectory(rainbow colored line) length is $26.48m$, planning time consuming is $59.89ms$.  \label{fig:outdoor}}{
		\includegraphics[width=0.9\columnwidth]{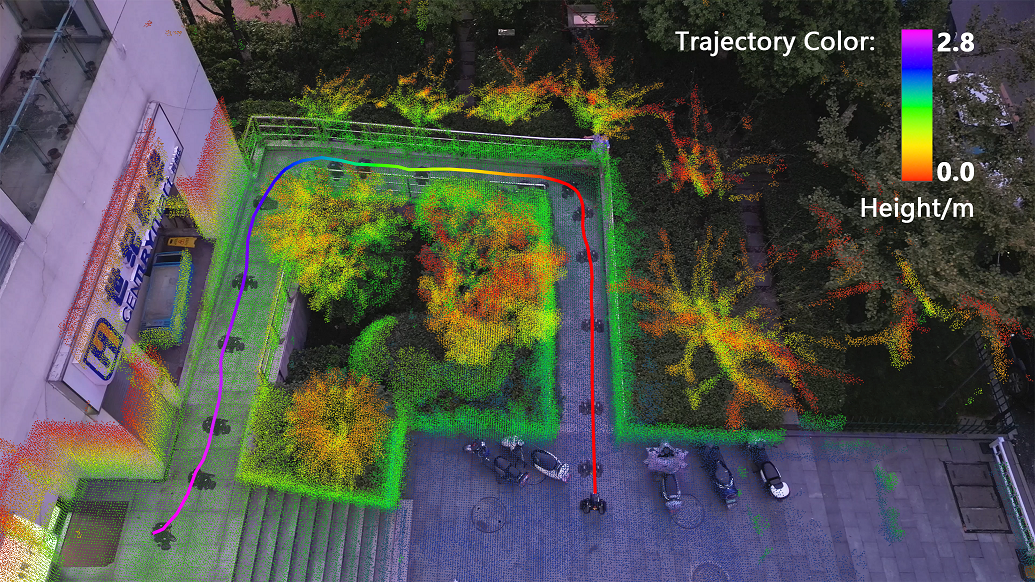}}
	\caption{Real-World Experiments}
	
\end{figure}

We present several experiments in both indoor and outdoor cluttered environments. In the indoor experiment, as shown in  Fig.\ref{fig:indoor}, we deliberately keep Diablo at a high altitude to test whether the added z-axis degrees of freedom could help the planner cope with such complex single-level indoor scenarios, thus Diablo needs to crouch down and pass under the table to reach the target.

\begin{figure}	
		\centering
		\includegraphics[width=0.88\columnwidth]{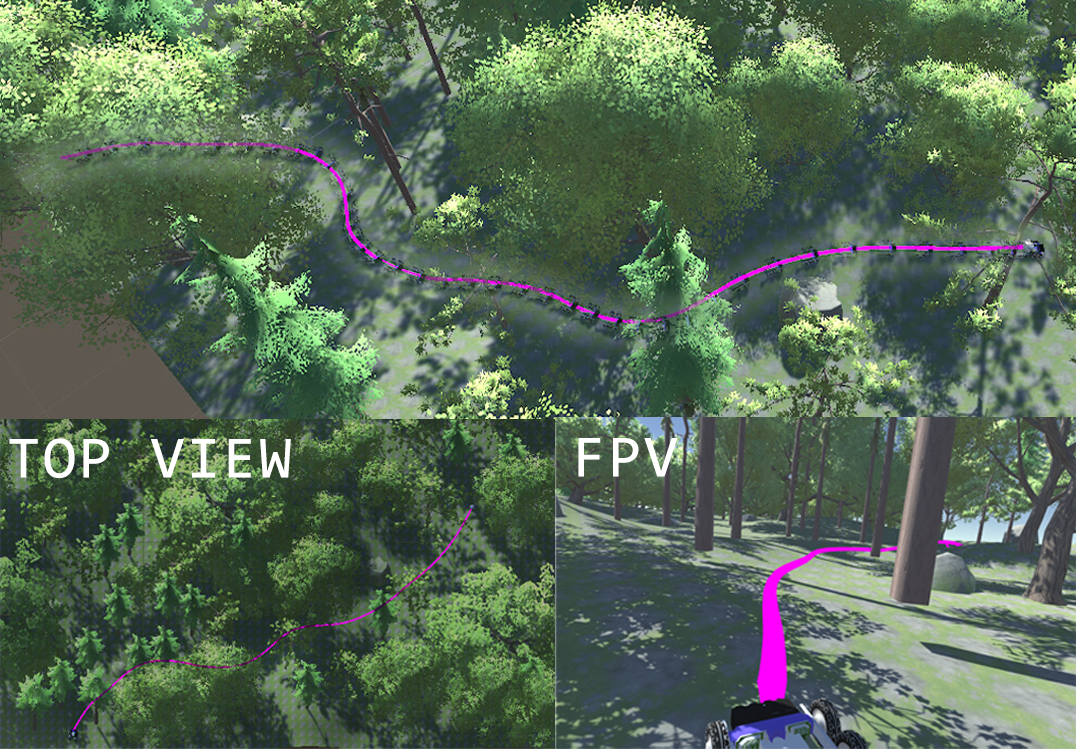}
		\caption{A four-wheeled differential robot navigating autonomously through the uneven forest.}\label{fig:uneven-forest}	
		\vspace{0.3cm}
	\vspace{-0.8cm}
\end{figure}

In the outdoor environment, Diablo is demanded to come up to a higher plane, as demonstrated in Fig.\ref{fig:outdoor}. There are two ways for humans to ascend this plane, by stairs or through barrier-free access, while for Diablo, there is only one way of the latter. In this test, the point clouds at the stairs are filtered out by the VGF because their normals are too horizontal, allowing the planner to ``intelligently'' select barrier-free access to generate trajectories. This test also proves that the proposed algorithm can efficiently solve the problems of motion planning in a multi-layered environment.

\subsection{Simulation Experiment }
\label{subsec:Simulation Experiment}
In order to testify the effectiveness of our method in more extensive, rugged and complex environments, we built an open-source simulation environment based on Unity\footnote{\url{https://unity.com/}}. There are %more than three of what we consider 
some typical scenarios, including uneven forests, crisscrossing viaducts, multi-level underground parking garages, etc. 
We simulate a four-wheeled differential robot driving in different environments. Similar to real-world experiments, we use the MPC controller mentioned in Sec.\ref{subsec:Real-World Experiments} for trajectory tracking. Fig.\ref{fig:uneven-forest} shows the robot navigating autonomously through one of the simulation scenarios — the uneven forest. More details about the simulation environment are given in the attached multimedia.

\begin{table}[t]
    \vspace{-0.8cm}
	\small
	\centering
	\renewcommand\arraystretch{1.2}
	\caption{\label{tab:table1} Methods Qualitative Comparison }
	\begin{tabular}{c|cccc}
		\hline
		\multirow{2}{*}{Method}& \multicolumn{1}{c}{Z-axis} & \multicolumn{1}{c}{Velocity} & 
		\multirow{2}{*}{Continuity} \\ 
	    &	\multicolumn{1}{c}{freedom} & \multicolumn{1}{c}{information}
	    
		\\ \hline
		\multirow{1}{*}{\begin{tabular}[c]{@{}c@{}}Proposed\end{tabular}}
		& \textbf{Support}                
		& \textcolor[RGB]{61,145,64}{\ding{52}}    &      \textbf{ 4-order }  \\ \hline
		\multirow{1}{*}{\begin{tabular}[c]{@{}c@{}}Liu's\cite{liu2015robotic}\end{tabular}}
		& No support 	&\textcolor[RGB]{227,23,13}{\ding{56}} &   0-order 
		\\ \hline
		\multirow{1}{*}{\begin{tabular}[c]{@{}c@{}}Kr{\"u}si's\cite{krusi2017driving} \end{tabular}} 
		& No support          
	    &\textcolor[RGB]{227,23,13}{\ding{56}}	& \textbf{ 4-order }  \\ \hline
	\end{tabular}
    \vspace{0.2cm}
\end{table}

\subsection{Benchmark Comparisons}
\label{subsec:Benchmark Comparisons}
\begin{figure}
	\centering
	\includegraphics[width=0.88\columnwidth]{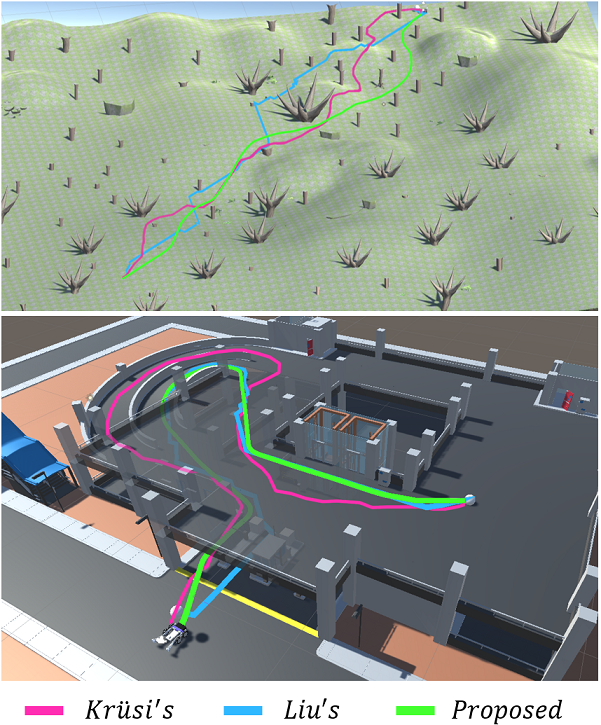}
	\caption{Trajectory visualization in uneven forest (top) and abandoned station (bottom).}\label{fig:simulation}
	\vspace{-0.4cm}
\end{figure}

In this section, we compare the proposed planning algorithm with Liu's method\cite{liu2015robotic} and Kr{\"u}si's method\cite{krusi2017driving}. We first qualitatively compared such methods and the results are in Table \ref{tab:table1}. It's clear that only our method support z-axis freedom, which is why Diablo could squat through the table in the indoor experiment. Besides, although Kr{\"u}si's method can generate $4^{th}$ order continuous trajectory, due to the fact that the trajectory is parameterized by the mileage, no information about the robot velocity and acceleration on the trajectory can be obtained. Liu's method only connects points on the point cloud without generating new waypoints, resulting in poor continuity and smoothness of the trajectory, as is evident in the comparison of the mean curvature mentioned in the next paragraph and the visualization of the trajectory in Fig.\ref{fig:simulation}.

\begin{table}[t]
    \vspace{-0.4cm}
	\small
	\centering
	\renewcommand\arraystretch{1.2}
	\caption{\label{tab:table2} Methods Quantitative Comparison }
	\begin{tabular}{c|clllllllll}
		\hline
		\multirow{2}{*}{Scenario}& \multirow{2}{*}{Method} &\multirow{2}{*}{$t_{a}(s)$} &\multirow{2}{*}{$len(m)$} &\multirow{2}{*}{$\kappa_m(m^{-1})$}\\
		%&\multicolumn{1}{c}{$t_{a}(s)$} &  \multicolumn{1}{c}{$len(m)$} &\multicolumn{1}{c}{$\kappa_m(m^{-1})$}\\
		&                         & \multicolumn{1}{c}{ } & \multicolumn{1}{c}{ } & \multicolumn{1}{c}{ }\\ \hline
		\multirow{3}{*}{\begin{tabular}[c]{@{}c@{}}Uneven\\Forest\end{tabular}}    
		& Proposed                
		&      \textbf{0.27}    &  \textbf{85.75} &  \textbf{0.042}\\
		& Liu's\cite{liu2015robotic}                 
		&      2.43 & 86.90 & 0.174\\
		& Kr{\"u}si's\cite{krusi2017driving}          
	    &   0.92	& 90.69  & 0.136 \\ \hline
		\multirow{3}{*}{\begin{tabular}[c]{@{}c@{}}Abandoned\\Station\end{tabular}} 
		& Proposed                
		&        \textbf{1.28} &  \textbf{62.75} &  \textbf{0.076}\\
		& Liu's\cite{liu2015robotic}               
		&         50.7  &  72.90 & 0.252 \\
		& Kr{\"u}si's\cite{krusi2017driving}           
		&   2.9     &     86.33                    &      0.104\\ \hline
	\end{tabular}
	\vspace{-0.4cm}
\end{table}

Further, to quantitatively measure the performance of the trajectory, some generic quantitative evaluation metrics are used to comparing, including terrain assessment time consuming $t_{a}$, trajectory length $len$ and mean curvature of the trajectory $\kappa_m$ which reflects the smoothness of the trajectory. We chose two scenes in our simulation environment, one is the uneven forest mentioned in Sec.\ref{subsec:Simulation Experiment}, and the other is an abandoned station with multi-layer structures, as shown in Fig.\ref{fig:simulation}. All parameters are finely tuned for the best performance of each compared method.

% \begin{figure}[t]  
% 		\vspace{-0.0cm}  
% 		\centering
% 		{\includegraphics[width=1.0\columnwidth]{FlightViz.png}}
% 		\caption{Trajectory visualization in uneven forest (top) and abandoned station (bottom).}\label{fig:simulation}
% 	\end{figure}
	
% \begin{figure}	
% 		\centering
% 		\includegraphics[width=1.0\columnwidth]{data_tl.png}
% 		\caption{Benchmark on computation efficiency}\label{fig:timecomp}	
% 		\vspace{0.3cm}
% \end{figure}

More than two thousand comparison tests are performed in each case with random starting and ending states. The result is summarized in Table \ref{tab:table2}. It states that the terrain assessment time of Liu's method is much longer than that of the other two. Despite the acceleration by parallel computing with GPUs, it takes a long time to transfer a large number of point clouds between GPUs and CPUs, and the large number of nearest neighbor searches required to construct the connectivity graph with the concept of k-NN also causes the terrain assessment in his method to be extremely time-consuming. Kr{\"u}si's method uses RRT-Connect as the front end, but in the optimization based on RRT*\cite{rrt-star}, the termination condition that takes into account the efficiency leads to insufficient iterations of the algorithm, thus there is a higher probability that the final trajectory is not topologically optimal, resulting in a longer trajectory length, as shown in the abandoned station scenario in Fig.\ref{fig:simulation}. % so that the final trajectory has a higher probability of not being topologically optimal globally, resulting in a longer trajectory length, as shown in the abandoned station scenario in Fig.\ref{fig:simulation}. 

In Table \ref{tab:table2}, our method achieves better performance in terms of terrain assessment time consuming $t_{a}$, trajectory length $len$ and mean curvature of the trajectory $\kappa_m$, which shows that the terrain assessment algorithm we propose is simpler and more effective, and the trajectories generated based on the proposed safety penalty field and planning framework achieve higher quality.

%Driving on pc rrt 搜出来初始轨迹之后,rewire过程是在原始轨迹上rand一个点,然后以一个范围d去搜索一个p_sample点,然后在树上找离这个p_sample最近的一个点,朝着这个p_sample方向走固定的步长找到一个新的点p_new,再以这个p_new的点为中心,r为半径搜索rrt_tree上的已有的节点,在这些节点里选择可以和这个p_new通过SRT,cubic轨迹相连的所有节点里距离最短的那个,因为他是拿rrt作为先验信息,而且为了兼顾效率他的rrt*没有迭代充分,这导致他更趋向于和先验的rrt保持拓扑一致,难以搜到一个更近的拓扑,导致他轨迹很长

\section{Conclusion}
\label{sec:Conclusion}
In this paper, we presented a construction method of a field function defined in $\mathbb{R}^3$ to constrain the motion of ground robot. Using the field function, an optimization-based planning algorithm is proposed for ground robots beyond 2D environment. Real-world experiments and simulation benchmark comparisons validate the efficiency and quality of our method. However, there is still space for improvement for our method. 
Although generating trajectories beyond 2D environment, the 3D grid map is still somewhat redundant, causing additional memory usage.
Moreover, the offline processing limits some online applications.

In the future, our research will focus on the improvement of memory utilization 
while ensuring the navigation capability beyond 2D environment for ground robots.
Besides, to further exploit the advantages of the efficiency of our algorithm, local re-planning will be taken into consideration and adapted to the dynamic environments.

\newlength{\bibitemsep}\setlength{\bibitemsep}{0.00\baselineskip}
\newlength{\bibparskip}\setlength{\bibparskip}{0pt}
\let\oldthebibliography\thebibliography
\renewcommand\thebibliography[1]{
    \oldthebibliography{#1}
    \setlength{\parskip}{\bibitemsep}
    \setlength{\itemsep}{\bibparskip}
}
\bibliography{references}

\end{document}